\documentclass{article}
\usepackage{spconf,amsmath,graphicx}
\usepackage{multirow}
\usepackage{array}
\usepackage{enumitem}
\setlist{nosep, leftmargin=14pt}
\usepackage{caption}
\usepackage{mwe} % to get dummy images

\usepackage{graphicx}
\title{Exploring Genetic-histologic Relationships in Breast Cancer}
\name{Ruchi Chauhan$^{1,2}$, PK Vinod$^1$, CV Jawahar$^2$}
\address{$^1$Center for Computational Natural Sciences \& Bioinformatics (CCNSB) \\
$^2$Center for Visual Information Technology (CVIT) \\ International Institute of Information Technology, Hyderabad, India}
\begin{document}
\ninept
\maketitle
\begin{abstract}
The advent of digital pathology presents opportunities for computer vision for fast, accurate, and objective solutions for histopathological images and aid in knowledge discovery. This work uses deep learning to predict genomic biomarkers - \textsc{tp53}  mutation, \textsc{pik3ca} mutation, \textsc{er} status, \textsc{pr} status, \textsc{her2} status, and intrinsic subtypes, from breast cancer histopathology images. Furthermore, we attempt to understand the underlying morphology as to how these genomic biomarkers manifest in images. Since gene sequencing is expensive, not always available, or even feasible, predicting these biomarkers from images would help in diagnosis, prognosis, and effective treatment planning. We outperform the existing works with a minimum improvement of 0.02 and a maximum of 0.13 \textsc{auroc} scores across all tasks. We also gain insights that can serve as hypotheses for further experimentations, including the presence of lymphocytes and karyorrhexis. Moreover, our fully automated workflow can be extended to other tasks across other cancer subtypes.
\end{abstract}


\section{Introduction}
\vspace{-1.5ex}
\label{sec:intro}
Histopathological evaluation involving microscopic examination of Hematoxylin \& Eosin \textsc{(h\&e)} stained specimen on the glass slide is considered a gold standard for cancer diagnosis. However, the manual assessment may be subjected to error, human bias, inter-intra pathologist variability, and low throughput. There have been remarkable advances through deep learning in cancer detection, mitosis detection \cite{mitosisLatest}, cancer metastasis detection \cite{metastasisLatest}, etc. These works focus on developing automated, fast, and accurate solutions for analysis routinely performed by pathologists. 

There have been significant developments in individualized diagnosis, prognosis, and treatment planning based on genomic biomarkers. However, despite the plummeting cost of genome sequencing, it can still be inaccessible, time-consuming, expensive, or infeasible due to the tissue being insufficient for excision. The association of histopathology findings from Whole Slide Images \textsc{(wsi)} and genomic alterations remains mostly unknown in different cancers. It is based on the hypothesis that image features encode the tumor’s underlying genotype. It is a challenging problem since genetic changes can manifest as subtle patterns in the images that are undetectable in an unaided approach to histopathology. Deep learning has shown promise in this aspect \cite{digPath, SOTAtp53pikrca}. 

We endeavor to improve classification and get insights into morphological features associated with genomic biomarkers.  We seek to explore the histological influence of mutation on the nuclei shape and size at the cellular level vs. its impact on the tumor microenvironment involving spatial aspects. Moreover, we examine any visual differences in terms of staining that could potentially be used by pathologists.

This work focuses on breast cancer, the most common cancer in women worldwide. Breast cancer is characterized by molecular features,  therapeutic responses,  disease progression, and preferential organ sites of metastases \cite{heterogeneous1}. Cancer can be caused by unrepaired alterations known as a mutation in the \textsc{dna} sequences that encode for genes. These mutations occur due to \textsc{dna} replication errors or environmental factors. \textsc{tp53} is a tumor suppressor gene that regulates uncontrolled growth and cell division. It is mutated across cancer types and is an independent risk factor for determining survival. Mutations in \textsc{pik3ca} oncogene leads to increased signaling for cell proliferation, which may result in a tumor. Further, the presence of Estrogen Receptor \textsc{(er)} and Progesterone Receptor \textsc{(pr)} is examined in the cancer cells. These hormone biomarkers \cite{ERPRstatus} are predictive of the efficacy of hormone therapy- the treatment strategy of modifying the tumor's hormonal milieu. \textsc{her2} proto-oncogene encodes for a growth-promoting protein in the breast cells whose overexpression may lead to a tumor. Such \textsc{her2} positive cases respond to therapies targetting \textsc{her2} protein. 

There has been a rising interest in classifying histopathology images with biomarkers in lung cancer, bladder cancer, prostate cancer, and breast cancer \cite{prostrateCancer,lungCancer,bladderCancer,SOTAerprher2,SOTAsubtype}. While these works establish that phenotype is predictive of the genotypic features using deep learning, they do not explore the features that were or could be used for classification. 

\begin{figure*}[t]
    \centering
    \includegraphics[width=\textwidth]{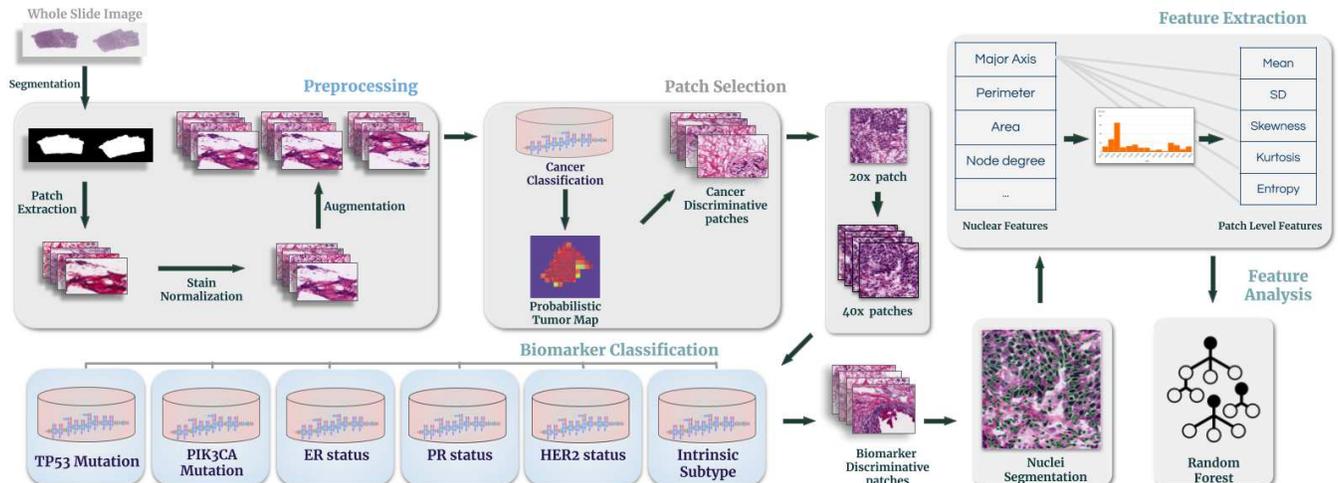}
    \captionsetup{belowskip=-1.5em, aboveskip=-0.25em}
    \caption{\textbf{Workflow: }The whole slide image is segmented to identify the tissue regions and divided into patches. The patches undergo colour normalization and augmentation to be used for cancer detection. The discriminative patches are converted to a higher resolution and used for biomarker prediction. The biomarker discriminative patches are analyzed by extracting features at the nuclei level and aggregated at the patch level. The features are tested using statistical analysis and random forest classifier.}
    \label{fig:my_pipeline}
\end{figure*}

\vspace{-2ex}
\section{Methods}
\vspace{-1.5ex}
\label{sec:format}

This work classifies the WSIs according to six biomarkers: mutations in \textsc{tp53, pik3ca, er} status, \textsc{pr} status, \textsc{her2} status, and intrinsic subtypes - Basal vs. non-Basal. The discriminative patches from these classifications were further analyzed to understand their distinguishing features in terms of intensity, morphology, spatial arrangement, and cell types. Random forest classification, nuclei annotations, and statistical analysis on gene expressions were used for this purpose. Fig \ref{fig:my_pipeline} shows the workflow in detail.

\textbf{Data Preparation:} The data used in this work is taken from \textsc{tp53} \cite{GDC}, which has a repository of 1054 anonymized \textsc{wsi} of breast invasive carcinoma patients with their genomic, pathologic, and de-identified clinical information. Images with highlighters or other artifacts were excluded, resulting in 708 cancer and 100 normal slides. For patients with multiple slides, the last biopsy slide was used. The standard practice of patch extraction resulted in an average of 3,000 patches per \textsc{wsi} of size $512\times512$ pixels at $20\times$ resolution. Color normalization using \cite{vahadane2016structure} accounted for the variations due to staining reagents and scanners by different manufacturers, protocols for slide preparation, etc.

\textbf{Classification:} Our work used InceptionNet-v3 pretrained on ImageNet, which by design, processes the images using multiple kernel filter sizes. Such architecture helps process pathology images from multiple fields of view, capturing the cellular along with glandular structures, befitting for our tasks. To handle the class imbalance (18-48\% in our dataset), we used random undersampling (with replacement) followed by random augmentation. Once a patch was sampled, a random augmentation was chosen with a random value and then inputted into the model. This approach introduced ample variation in the training dataset without multiplying its size and provided the desired robustness at the expense of convergence time. Colour normalization with stain augmentation is proven effective by a systematic analysis \cite{stainAug}. We used light \textsc{hsv} transformation, \textsc{hed} transformation, additive Gaussian noise, flip, and rotations. Test-time augmentation (without randomness) and mixed-precision training were used. \footnote{All implementation details \& data summary can be found at: https://github.com/theRuchiChauhan. Codes will be made available upon publication.}

It must be noted that the labels for the data are available at the patient level, i.e., for slides, and may not be true for the individual patches. The tumor within a slide may be localized, and not all patches from a slide may be cancerous containing the genomic information.  Hence there is a mislabelling in the training set that can deteriorate the performance. To overcome this, we identify diagnostically salient regions on the \textsc{wsi}. To keep our workflow fully automated, we obtain machine-generated annotations employing deep learning instead of pathologists' annotations. Moreover, manual annotation on a slide is not viable for mutations as the pathologists do not look for mutations in the histopathological images.
To that end, cancer detection - a binary classification of cancer versus normal patches was used to find the \textit{discriminative patches}. This task used all the available normal (non-cancer) slides and an equal number of Cancer slides. The remaining cancer slides were kept as an external test set, later used for biomarker prediction. The model gets uncorrupted labels from the normal slides and can understand a non-cancer patch. This seems to be sufficient even though the problem mentioned above persists for the cancer slides, i.e., not all cancer slide patches are cancerous. The cancer detection is done at 20$\times$ resolution, and the discriminative patches are translated to 40$\times$ resolution, the highest magnification, to capture finer level details. This translation is done by zooming using Lanczos interpolation, and cropping giving four 40$\times$ patches for every 20$\times$ patch. These discriminative patches are used for the main tasks of genomic biomarker prediction. Roughly 150,000 patches were obtained at 40$\times$ resolution.

\textit{Discriminative patches} are the patches that were correctly classified by the model with high confidence. The softmax probabilities of the predictions from the classification model were used to obtain the confidence scores. Despite achieving better performance, modern neural networks tend to be overconfident. This can be attributed to the increase in width and depth compared to the older neural networks like LeNet, and methods like batch normalization and weight decay. Thus, the probabilities of prediction are not representative of the true correctness likelihood and calls for a calibrated confidence score. To this end, we use a straightforward calibration technique: Temperature Scaling \cite{confidenceCalibration}. The confidence score is calculated as
\begin{equation}
\hat{q} = \sigma_{SM}(z/T)
\end{equation}
 where $\sigma_{SM}$ is softmax operation, z refers to the logits, and T is the temperature computed by logistic regression.
These confidence scores were thresholded at 0.9, above which the patch is deemed `discriminative'. Filtering discriminative patches from the cancer detection model reduced the dataset to roughly 45\%. These patches are termed as Cancer Discriminative \textsc{(cd)} patches and used for biomarker classification tasks. The discriminative patches from each biomarker classification model are called Biomarker Discriminative \textsc{(bd)} patches.   

\begin{figure*}[t]
    \centering
    \includegraphics[width=0.9\textwidth]{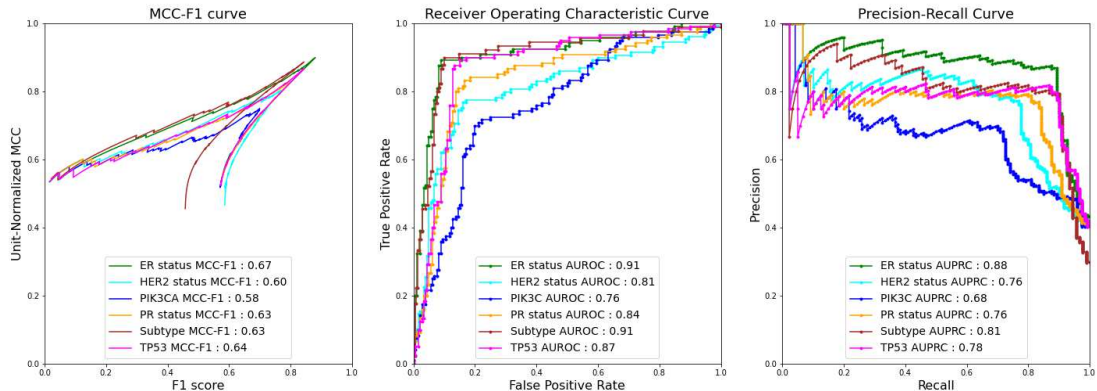}
    \captionsetup{belowskip=-1.5em, aboveskip=-0.25em}
    \caption{Classification Results of Genomic Biomarkers using InceptionNet. [L-R] \textsc{mcc-f1} curve, Receiver Operating Characteristic Curve, Precision-Recall Curve. Overall, \textsc{er} performed the best, while \textsc{pik3ca} performed the worst}
    \label{fig:my_results}
\end{figure*}  

%Nuclei Segmentation\\
 \textbf{Feature Analysis: }Nuclei segmentation \cite{NucSegphoulady} followed by ellipse fitting on each nucleus was used to examine the discriminative patches. 
For each of the nuclei, the following features were computed:
\textit{Morphological features: }minor axis, major axis, ratio of major axis to minor axis, area, perimeter, circularity, eccentricity, and solidity. \textit{Intensity Features: } mean pixel value for red, green, and blue channels. Note that the patches are colour normalized.
\textit{Spatial Features:} For each patch, a Delaunay triangulation graph connecting the centroids of the nuclei as nodes were constructed. From this graph, the following were calculated for each nucleus: minimum distance from a neighbor, the maximum distance from a neighbor, the average of distances with all neighbors, and the number of neighbors (node degree). All these features calculated for nuclei were aggregated at the patch level. The distribution statistics: mean, standard deviation, entropy, skewness, and kurtosis were taken for each patch using a ten-binned histogram. The histogram was $L1$ normalized to mitigate the effects of an unequal number of nuclei across patches. Thus, we obtained a total of 75 features.

To analyze the features characteristic of a biomarker class, we classify the patches using Random Forest. To further explore the relative importance of features, an ablation study was performed using combinations of features. The classification is done on biomarker discriminative patches and also on the cancer discriminative patches for each task. The complete pipeline is shown in Fig. \ref{fig:my_pipeline}. To further explore the tumor microenvironment at the cellular level, nuclei annotation was done using the tool provided by \cite{stainTool}, which employs mask-\textsc{rcnn} for nuclei segmentation. 

\vspace{-2ex}
\section{Results and discussion}
\vspace{-1.5ex}

\begin{figure}[ht]
    \centering
    \includegraphics[width=\linewidth]{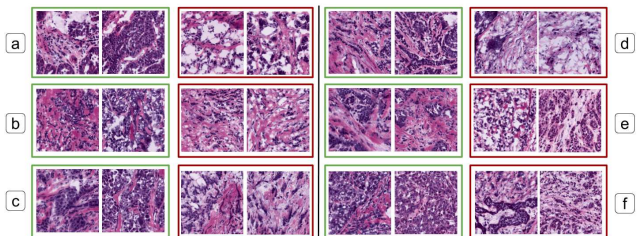}
    \captionsetup{belowskip=-1.5em, aboveskip=-0.15em}
    \caption{Top Biomarker Discriminative patches. a: \textsc{er} status, b: \textsc{pr} status, c: \textsc{her2} status, d: \textsc{tp53}, e: \textsc{pik3ca}, f: Intrinsic Subtype. Green Box: positive/mutated/Basal subtype, Red Box: negative/not-mutated/non-basal subtypes}
    \label{fig:my_Qualresults}
\end{figure}

The model achieved slide level ~0.99 \textsc{auroc} for Cancer vs. Normal classification, comparable to that reported in the literature on the \textsc{tcga} dataset \cite{CanAcc}. 
Our models outperform the results reported elsewhere in the literature for biomarker prediction Table \ref{TableDL}. The slide level results are calculated by aggregation using the average of probabilities. We observed an expected improvement in the biomarker classification by training only on our cancerous discriminative patches. The baseline approach used all the patches from the \textsc{wsis}.  \textsc{auroc} has been tabulated for benchmarking. Additional metrics - \textsc{auprc, mcc-f1} are shown in Fig. \ref{fig:my_results}. The \textsc{mcc-f1} curve reported is more appropriate for unbalanced datasets than other metrics as it provides a complete summary of the confusion matrix \cite{MCC}. Qualitative results are shown in Fig. \ref{fig:my_Qualresults}

%##################################

\begin{table}[h!]
\captionsetup{ aboveskip=-0.05em}
\caption{\label{TableDL}Classification Results of Genomic Biomarkers using InceptionNet: 
 $^\dagger$ : \cite{SOTAtp53pikrca}, $^\ddagger$ : \cite{SOTAerprher2},  \textsuperscript{§} : \cite{SOTAsubtype}}
\centering
\begin{tabular}{|p{0.81cm}|p{0.5cm}|p{0.5cm}|p{0.76cm}|p{0.5cm}|p{0.5cm}|p{0.6cm}|p{0.85cm}|}
\hline 
{\scriptsize{}AUROC} & {\footnotesize{}Level} & {\footnotesize{}TP53} & {\footnotesize{}PIK3CA} & {\footnotesize{}ER} & {\footnotesize{}PR}\textsuperscript{} & {\footnotesize{}HER2}\textsuperscript{} & {\footnotesize{}Subtype}\tabularnewline
\hline 
\hline 
\multirow{2}{*}{{\footnotesize{}Ours}} & {\footnotesize{}patch} & {\footnotesize{}0.829} & {\footnotesize{}0.721} & {\footnotesize{}0.866} & {\footnotesize{}0.820} & {\footnotesize{}0.798} & {\footnotesize{}0.877}\tabularnewline
\cline{2-8} 
 & {\footnotesize{}slide} & \textbf{\footnotesize{}0.875} & \textbf{\footnotesize{}0.765} & \textbf{\footnotesize{}0.910} & \textbf{\footnotesize{}0.839} & \textbf{\footnotesize{}0.811} & \textbf{\footnotesize{}0.909}\tabularnewline
\hline 
\multirow{2}{*}{{\footnotesize{}Baseline}} & {\footnotesize{}patch} & {\footnotesize{}0.677} & {\footnotesize{}0.565} & {\footnotesize{}0.665} & {\footnotesize{}0.614} & {\footnotesize{}0.666} & {\footnotesize{}0.703}\tabularnewline
\cline{2-8} 
 & {\footnotesize{}slide} & {\footnotesize{}0.643} & {\footnotesize{}0.541} & {\footnotesize{}0.632} & {\footnotesize{}0.578} & {\footnotesize{}0.622} & {\footnotesize{}0.685}\tabularnewline
\hline 
{\footnotesize{}Best} & {\footnotesize{}slide} & {\footnotesize{}0.75\dag{}} & {\footnotesize{}0.63\dag{}} & {\footnotesize{}0.89}\textsuperscript{{\footnotesize{}\ddag{}}} & {\footnotesize{}0.81}\textsuperscript{{\footnotesize{}\ddag{}}} & {\footnotesize{}0.79}\textsuperscript{{\footnotesize{}\ddag{}}}{\footnotesize{} } & {\footnotesize{}0.826}\textsuperscript{{\footnotesize{}§}}\tabularnewline
\hline 
\end{tabular}
\end{table}
%##################################
%%%%%%%%%%%%%%%%%%%%%%%%%%%%%%%%%%%%%%%%%%% RF combi table
\begin{table*}[h!]
\centering
\captionsetup{belowskip=-2em, aboveskip=-0.15em}
\caption{\label{RFtable}Results from random forest classifier on Biomarker Discriminative \textsc{(bd)} \& Cancer Discriminative \textsc{(cd)} Patches. \\
Best performance in bold and next best underlined. }
% \vspace{2px}
\begin{tabular}{|c|c|c|c|c|c|c|c|c|c|c|c|c|}
\hline 
\multirow{2}{*}{
{\footnotesize{}AUPRC}
} & \multicolumn{2}{c|}{{\footnotesize{}TP53}} & \multicolumn{2}{c|}{{\footnotesize{}PIK3CA}} & \multicolumn{2}{c|}{{\footnotesize{}ER}} & \multicolumn{2}{c|}{{\footnotesize{}PR}} & \multicolumn{2}{c|}{{\footnotesize{}HER2}} & \multicolumn{2}{c|}{{\footnotesize{}Subtype}}\tabularnewline
\cline{2-13} 
 & {\footnotesize{}BD} & {\footnotesize{}CD} & {\footnotesize{}BD} & {\footnotesize{}CD} & {\footnotesize{}BD} & {\footnotesize{}CD} & {\footnotesize{}BD} & {\footnotesize{}CD} & {\footnotesize{}BD} & {\footnotesize{}CD} & {\footnotesize{}BD} & {\footnotesize{}CD}\tabularnewline
\hline 
{\footnotesize{}All Features (n=75)} & \textbf{\footnotesize{}0.934} & \textbf{\footnotesize{}0.657} & {\footnotesize{}\underline{0.882}} & \textbf{\footnotesize{}0.618} & \textbf{\footnotesize{}0.837} & \textbf{\footnotesize{}0.655} & {\footnotesize{}\underline{0.862}} & {\footnotesize{}\underline{0.640}} & {\footnotesize{}\underline{0.927}} & {\footnotesize{}\underline{0.584}} & \textbf{\footnotesize{}0.828} & \textbf{\footnotesize{}0.696}\tabularnewline
\hline 
{\footnotesize{}Intensity Features (n=15)} & {\footnotesize{}0.877} & {\footnotesize{}0.611} & {\footnotesize{}0.821} & {\footnotesize{}0.585} & {\footnotesize{}0.733} & {\footnotesize{}0.613} & {\footnotesize{}0.729} & {\footnotesize{}0.587} & {\footnotesize{}0.865} & {\footnotesize{}0.569} & {\footnotesize{}0.763} & {\footnotesize{}0.661}\tabularnewline
\hline 
{\footnotesize{}Spatial Features (n=20)} & {\footnotesize{}0.838} & {\footnotesize{}0.592} & {\footnotesize{}0.741} & {\footnotesize{}0.571} & {\footnotesize{}0.674} & {\footnotesize{}0.590} & {\footnotesize{}0.681} & {\footnotesize{}0.577} & {\footnotesize{}0.799} & {\footnotesize{}0.541} & {\footnotesize{}0.690} & {\footnotesize{}0.629}\tabularnewline
\hline 
{\footnotesize{}Morphology Features (n=40)} & {\footnotesize{}0.799} & {\footnotesize{}0.581} & {\footnotesize{}0.828} & {\footnotesize{}0.605} & {\footnotesize{}0.74} & {\footnotesize{}0.599} & {\footnotesize{}0.832} & {\footnotesize{}0.612} & {\footnotesize{}0.909} & {\footnotesize{}0.558} & {\footnotesize{}0.742} & {\footnotesize{}0.635}\tabularnewline
\hline 
{\footnotesize{}Morphology + Spatial Features (n=60)} & {\footnotesize{}0.866} & {\footnotesize{}0.623} & {\footnotesize{}0.851} & {\footnotesize{}0.603} & {\footnotesize{}0.768} & {\footnotesize{}0.622} & {\footnotesize{}0.845} & {\footnotesize{}0.613} & {\footnotesize{}0.904} & {\footnotesize{}0.561} & {\footnotesize{}0.765} & {\footnotesize{}0.662}\tabularnewline
\hline 
{\footnotesize{}Spatial + Intensity Features (n=35)} & {\footnotesize{}\underline{0.933}} & {\footnotesize{}\underline{0.654}} & {\footnotesize{}0.840} & {\footnotesize{}0.601} & {\footnotesize{}0.788} & {\footnotesize{}0.640} & {\footnotesize{}0.773} & {\footnotesize{}0.601} & {\footnotesize{}0.891} & {\footnotesize{}0.555} & {\footnotesize{}0.797} & {\footnotesize{}0.676}\tabularnewline
\hline 
{\footnotesize{}Morphology + Intensity Features (n=55)} & {\footnotesize{}0.922} & {\footnotesize{}0.624} & \textbf{\footnotesize{}0.883} & {\footnotesize{}\underline{0.617}} & {\footnotesize{}\underline{0.820}} & {\footnotesize{}\underline{0.648}} & \textbf{\footnotesize{}0.864} & \textbf{\footnotesize{}0.642} & \textbf{\footnotesize{}0.929} & \textbf{\footnotesize{}0.588} & {\footnotesize{}\underline{0.825}} & {\footnotesize{}\underline{0.683}}\tabularnewline
\hline 
\end{tabular}
\end{table*}

Table \ref{RFtable} presents the results of the biomarker classification on biomarker and cancer discriminative patches using random forest. The same set of features did not work as well on all cancer discriminative patches compared to biomarker discriminative patches. This substantiates our claim that the generated patches are indeed discriminative of the genomic information. It might be possible that not all cancer regions contain the manifestation of biomarkers. 

We observe that for \textsc{tp53}, the intensity features alone outperform morphological and spatial features combined. Moreover, intensity and spatial features together perform comparably to all features taken together. This contrasts with other biomarkers, where the spatial features do not seem to be contributing much. Morphology features worked reasonably well for almost all the tasks except \textsc{tp53}.  

\begin{figure}[ht]
    \centering
    \includegraphics[width=0.85\linewidth]{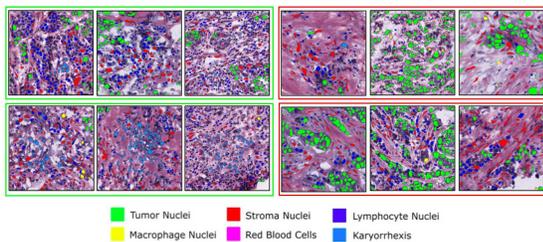}
    \captionsetup{belowskip=-1.5em, aboveskip=-0.05em}
    \caption{Prevalence of lymphocytic nuclei in \textsc{tp53} mutated [Top-left, green] vs \textsc{tp53} non-mutated [Top-right, red] patches. Presence of Karyorrhexis in \textsc{pik3ca} mutated [Bottom-left, green] vs \textsc{pik3ca} non-mutated [Bottom-right, red] patches. Tumor nuclei can be observed in both \textsc{tp53} \& \textsc{pik3ca}.} 
    \label{fig:my_annotresults}
\end{figure}

\begin{figure}[h!]
    \centering
    \includegraphics[width=0.75\linewidth]{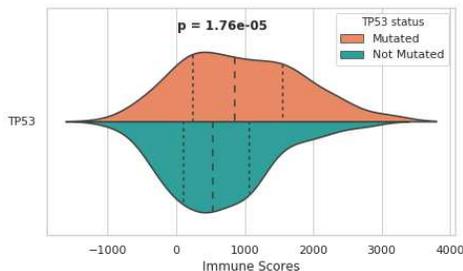}
    \captionsetup{belowskip=-1.5em, aboveskip=-0.05em}
    \caption{Immune Scores indicating immune infiltration of the \textsc{tp53} mutated gene expression samples are significantly higher than that of non-mutated genes. P-value from Mann Whitney U test.}
    \label{fig:my_violin}
\end{figure}

Interestingly, we observed a large number of lymphocyte nuclei in \textsc{tp53} mutated discriminative patches. Lymphocytes are the white blood cells whose presence indicates an immunological response. Further investigation using gene expression and immune scores obtained from \textsc{estimate} \cite{ESTIMATE} suggest the correlation between \textsc{tp53} mutation and increased immunologic activities. The immune scores signify the infiltration of immune cells in tumor tissues. A P-value of 1.76e-05 was obtained from the Mann Whitney U test between the immune scores of \textsc{tp53} mutated and nonmutated gene expression samples (Fig. \ref{fig:my_violin}). 
These observations go along with the demonstrated involvement of \textsc{tp53} in crucial aspects of tumor immunology and the homeostatic regulation of the immune responses \cite{TP53immune}.
Moreover, we found cells undergoing karyorrhexis, nuclei fragmentation during a stage of cell death, in \textsc{pik3ca} mutated discriminative patches, as shown in Fig. \ref{fig:my_annotresults}. 
\vspace{-2ex}
\section{Conclusion}
\vspace{-1.5ex}
\label{sec:majhead}

This work presented the classification of breast cancer genomic biomarkers from histopathological images. Conventionally, such classification is performed using gene expression data. Automated pipelines such as ours aim to augment pathological workflow while the pathologists may handle higher-level decisions. Despite the remarkable performance of deep learning solutions in computer-aided diagnosis, there is still legitimate skepticism for widespread clinical adoption. Hence, there is a need for understanding the classification done by a deep neural network into human interpretable features to help reduce the opacity of the black box models and generate knowledge. It is worth noting that the characteristic features are from the network's perspective and the biological significance derived herein warrants further validation.

\vspace{-2ex}
\section{Compliance with ethical standards}
\vspace{-1.8ex}
This study used data from \textsc{tcga} following the data access policies. Ethical approval was not required.
\vspace{-2ex}
\section{Acknowledgments}
\vspace{-1.8ex}
No funding was received for this work, and the authors have no relevant financial or non-financial interests to disclose.

\vspace{-2ex}
\bibliographystyle{IEEEbib}
{\footnotesize{
\bibliography{strings,refs}

\begin{thebibliography}{10}

\bibitem{mitosisLatest}
Saha et~al.,
\newblock ``Efficient deep learning model for mitosis detection using breast
  histopathology images,''
\newblock {\em Computerized Med. Imag. and Graph.}, vol. 64, pp. 29--40, 2018.

\bibitem{metastasisLatest}
Pham et~al.,
\newblock ``Detection of lung cancer lymph node metastases from whole-slide
  histopathologic images using a two-step deep learning approach,''
\newblock {\em Amer. J. pathology}, vol. 189, no. 12, pp. 2428--2439, 2019.

\bibitem{digPath}
Couture et~al.,
\newblock ``Image analysis with deep learning to predict breast cancer grade,
  er status, histologic subtype, and intrinsic subtype,''
\newblock {\em NPJ breast cancer}, vol. 4, no. 1, pp. 1--8, 2018.

\bibitem{SOTAtp53pikrca}
Kather et~al.,
\newblock ``Pan-cancer image-based detection of clinically actionable genetic
  alterations,''
\newblock {\em Nature Cancer}, pp. 1--11, 2020.

\bibitem{heterogeneous1}
et~al. Reis-Filho,
\newblock ``Gene expression profiling in breast cancer: classification,
  prognostication, and prediction,''
\newblock {\em The Lancet}, vol. 378, no. 9805, pp. 1812--1823, 2011.

\bibitem{ERPRstatus}
Dunnwald et~al.,
\newblock ``Hormone receptor status, tumor characteristics, and prognosis: a
  prospective cohort of breast cancer patients,''
\newblock {\em Breast cancer res.}, vol. 9, no. 1, pp. R6, 2007.

\bibitem{prostrateCancer}
Schaumberg et~al.,
\newblock ``H\&e-stained whole slide image deep learning predicts spop mutation
  state in prostate cancer,''
\newblock {\em BioRxiv}, p. 064279, 2018.

\bibitem{lungCancer}
Coudray et~al.,
\newblock ``Classification and mutation prediction from non--small cell lung
  cancer histopathology images using deep learning,''
\newblock {\em Nature med.}, vol. 24, no. 10, pp. 1559, 2018.

\bibitem{bladderCancer}
Xu~et~al.,
\newblock ``Using transfer learning on whole slide images to predict tumor
  mutational burden in bladder cancer patients,''
\newblock {\em bioRxiv}, p. 554527, 2019.

\bibitem{SOTAerprher2}
Rawat et~al.,
\newblock ``Deep learned tissue “fingerprints” classify breast cancers by
  er/pr/her2 status from h\&e images,''
\newblock {\em Scientific Reports}, vol. 10, no. 1, pp. 1--13, 2020.

\bibitem{SOTAsubtype}
Jaber et~al.,
\newblock ``A deep learning image-based intrinsic molecular subtype classifier
  of breast tumors reveals tumor heterogeneity that may affect survival,''
\newblock {\em Breast Cancer Research}, vol. 22, no. 1, pp. 12, 2020.

\bibitem{GDC}
Grossman et~al.,
\newblock ``Toward a shared vision for cancer genomic data,''
\newblock {\em New England J. of Med.}, vol. 375, no. 12, pp. 1109--1112, 2016.

\bibitem{vahadane2016structure}
Vahadane et~al.,
\newblock ``Structure-preserving color normalization and sparse stain
  separation for histological images,''
\newblock {\em IEEE Trans Med. Imag.}, vol. 35, no. 8, pp. 1962--1971, 2016.

\bibitem{stainAug}
Tellez et~al.,
\newblock ``Quantifying the effects of data augmentation and stain color
  normalization in convolutional neural networks for computational pathology,''
\newblock {\em Med. image anal.}, vol. 58, pp. 101544, 2019.

\bibitem{confidenceCalibration}
Guo et~al.,
\newblock ``On calibration of modern neural networks,''
\newblock {\em ICML 2017}, 2017.

\bibitem{NucSegphoulady}
Phoulady et~al.,
\newblock ``Nucleus segmentation in histology images with hierarchical
  multilevel thresholding,''
\newblock in {\em Med. Imag. 2016: Digit. Pathology}. Int. Soc. for Opt. and
  Photonics, 2016, vol. 9791, p. 979111.

\bibitem{stainTool}
Wang et~al.,
\newblock ``Computational staining of pathology images to study the tumor
  microenvironment in lung cancer,''
\newblock {\em Cancer Research}, vol. 80, no. 10, pp. 2056--2066, 2020.

\bibitem{CanAcc}
Noorbakhsh et. al,
\newblock ``Pan-cancer classifications of tumor histological images using deep
  learning,''
\newblock {\em bioRxiv}, p. 715656, 2019.

\bibitem{MCC}
Cao et~al.,
\newblock ``The mcc-f1 curve: a performance evaluation technique for binary
  classification,''
\newblock {\em arXiv preprint arXiv:2006.11278}, 2020.

\bibitem{ESTIMATE}
Wong et~al.,
\newblock ``Characterization of cytokinome landscape for clinical responses in
  human cancers,''
\newblock {\em Oncoimmunology}, vol. 5, no. 11, pp. e1214789, 2016.

\bibitem{TP53immune}
Mu{\~n}oz-Fontela et~al.,
\newblock ``Emerging roles of p53 and other tumour-suppressor genes in immune
  regulation,''
\newblock {\em Nature Reviews Immunology}, vol. 16, no. 12, pp. 741--750, 2016.

\end{thebibliography}
}}
\end{document}